\documentclass{article}

\usepackage[preprint]{neurips_2026}        

\usepackage[utf8]{inputenc} 
\usepackage[T1]{fontenc}    
\usepackage{url}            
\usepackage{nicefrac}       
\usepackage[table]{xcolor}         

\usepackage{microtype}
\usepackage{graphicx}
\usepackage{subcaption}
\usepackage{booktabs} 

\usepackage{hyperref}       

\usepackage{amsmath}
\usepackage{amssymb}
\usepackage{mathtools}
\usepackage{amsthm}
\usepackage{multirow}
\usepackage{makecell}
\usepackage{xspace}
\usepackage{bbding}      

\newcommand{\meanstd}[2]{#1$_{\pm#2}$}
\newcommand{\method}{URDF-Anything+\xspace}

\usepackage[capitalize,noabbrev]{cleveref}

\theoremstyle{plain}

\theoremstyle{definition}

\theoremstyle{remark}

\usepackage[textsize=tiny]{todonotes}
\usepackage{bm}
\newcommand{\z}{\bm{z}}
\newcommand{\bv}{\bm{v}}

\title{URDF-Anything+:  End-to-End Generation for Simulation-Ready Articulated Assets}

\author{
Zhuangzhe Wu\thanks{Equal contribution, \textsuperscript{\Envelope}Corresponding author.} \hspace{0.1mm} \textsuperscript{\rm 1} \hspace{0.1mm}
Yue Xin$^{*}$\textsuperscript{\rm 1} \hspace{0.1mm}
Chengkai Hou\textsuperscript{\rm 1} \hspace{0.1mm}
Minghao Chen\textsuperscript{\rm 2} \\
\textbf{Yaoxu Lyu}\textsuperscript{\rm 1} \hspace{0.1mm}
\textbf{Jieyu Zhang}\textsuperscript{\rm 3} \hspace{0.1mm}
\textbf{Shanghang Zhang\textsuperscript{\rm 1}~\textsuperscript{\Envelope}} \hspace{0.1mm} \\
\textsuperscript{\rm 1}Peking University \\
\textsuperscript{\rm 2}Visual Geometry Group, University of Oxford \\
\textsuperscript{\rm 3}University of Washington \vspace{0.2cm}\\
Project page: \url{https://urdf-anything-plus.github.io/}
}

\begin{document}

\maketitle

\begin{abstract}
Articulated objects are fundamental for robotics, simulation of physics, and interactive virtual environments. However, recovering them from visual observations is inherently challenging, as images provide only partial and ambiguous cues about both part geometry and their underlying kinematic structure. Existing approaches typically rely on multi-stage pipelines, retrieval from asset libraries, or explicit part segmentation.
We present \textit{\method}, an end-to-end autoregressive diffusion framework that generates simulation-ready URDF models directly from a single RGB image. Conditioned on visual observations and object geometry, \method operates in a structured latent space and jointly models part geometry and articulation in a unified generation process. Specifically, the model sequentially predicts each articulated part together with its associated joint parameters, while a termination token dynamically determines the number of parts. This design enables direct generation of fully executable URDFs without external retrieval or post-processing stages.
Experiments on large-scale articulated object benchmarks demonstrate that \method outperforms prior methods in geometric reconstruction quality, joint parameter estimation, and physical executability, while being substantially more efficient than existing multi-stage approaches. Furthermore, the generated URDFs serve as faithful digital twins, enabling the zero-shot transfer of manipulation policies trained purely in simulation.
\end{abstract}
\section{Introduction}%
\label{sec:intro}
Robotic manipulation, physics-based simulation, and interactive virtual environments rely on simulation-ready articulated object models~\cite{xiang2020sapien,mo2019partnet}. These models must capture both geometry and kinematic structure to support physical interaction and control~\cite{liu2025survey}. However, constructing such assets (e.g., in URDF or XML formats) remains largely manual and labor-intensive. Recovering articulated object models directly from visual observations is therefore a highly desirable but challenging problem. 

\begin{figure}[t]
  \centering
  \makebox[\textwidth][c]{\includegraphics[width=1.02\textwidth]{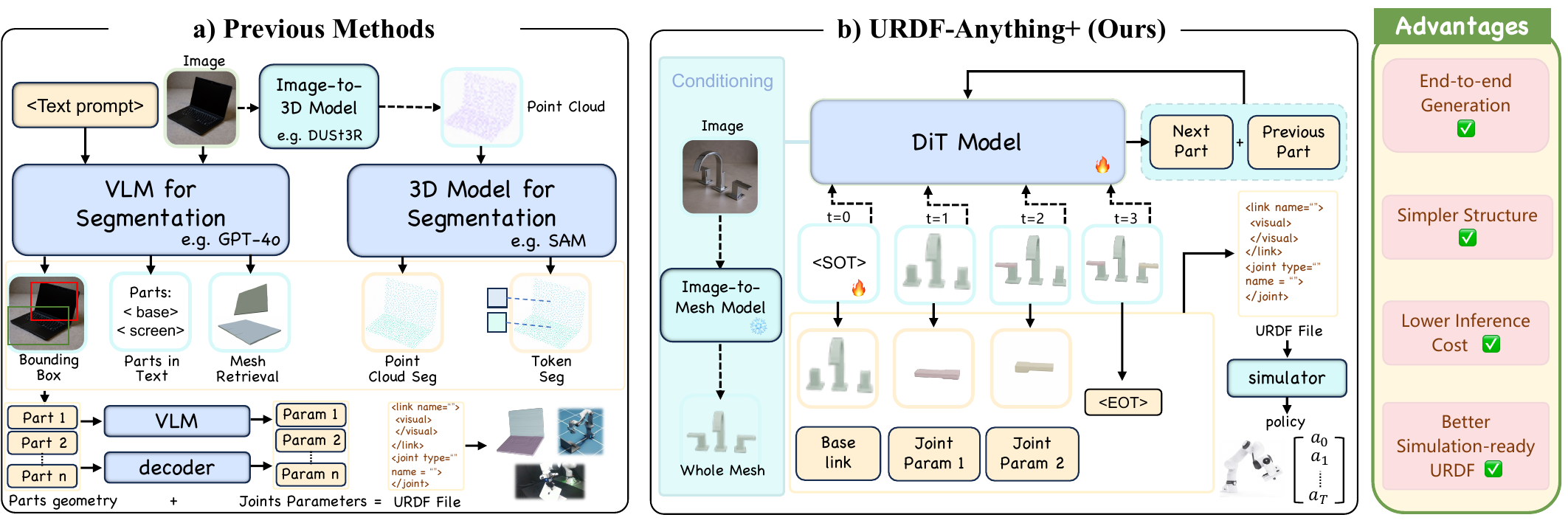}}
  \caption{\textbf{Comparison between previous methods and our method.}  In (a), we summarize previous generation paradigms for articulated objects, including different segmentation approaches and joint parameter generation methods. In (b), we briefly illustrate our method and its advantages.}
  \label{fig:model_compare}
\end{figure}

A model has to solve two coupled subproblems at once: it must decompose the object into rigid parts, and for every pair of connected parts it must recover the joint specification, including the joint origin, rotation axis, joint type, and motion limits. It requires jointly inferring part geometry and their underlying kinematic relationships from incomplete and ambiguous visual cues. Faced with this difficulty, prior work has largely converged on multi-stage pipelines that decouple segmentation and articulation, as illustrated in Fig.~\ref{fig:model_compare}(a). Some methods retrieve part meshes from an asset library and fit articulation on top~\cite{urdformer,le2025articulateanything}, bounding geometry to the library's coverage. Others delegate part segmentation to a vision-language model before synthesizing geometry and joints~\cite{le2025articulateanything,cao2025physx, simartdecomposingmonolithicmeshes}. Another line first segments part-level point clouds and then generates joint parameters with a separate code head~\cite{mandi2025realcode}. Such pipelines compound
error across stages or generalize poorly beyond categories well represented in retrieval libraries.


In this paper, we propose \textbf{\method}, an end-to-end framework for generating fully executable articulated URDF models directly from visual observations. Given an image and object-level 3D geometry, \method predicts both link geometries and their associated joint parameters in a part-by-part autoregressive manner, producing simulation-ready models that can be directly deployed in standard physics engines.

Our key insight is that articulated objects are inherently compositional and can be naturally modeled through autoregressive generation. By sequentially generating parts together with their kinematic relationships, \method captures geometry and articulation in a unified process, avoiding the need for multi-stage pipelines or external asset retrieval.

To achieve this, we adopt an autoregressive diffusion-based formulation that operates in a structured latent 3D space. At each step, the model predicts both part geometry and joint parameters, ensuring globally consistent shape and kinematics across the entire object. This design enables a unified perception-to-simulation pipeline, supporting scalable digital twin construction, free-form articulated object generation, and robotic learning.

We first evaluate \method on large-scale articulated object benchmarks, where our method significantly outperforms previous approaches in both geometric reconstruction and joint parameter prediction, and achieves a substantially higher rate of physically executable URDFs, even for challenging objects outside the training distribution. We further introduce \textbf{Real-Follow-Sim}, a deployment scheme that uses our generated URDFs as digital twins so that policies trained purely in simulation transfer zero-shot to real robots.


Our contributions are threefold:
\begin{enumerate}
\item We introduce \method, an end-to-end autoregressive diffusion framework that generates articulated URDF models directly from a single image in a single forward pass. Parts are synthesized in a unified coordinate system, simplifying URDF assembly. The model is lightweight and runs in $34.70$\,s on average, considerably faster than prior multi-stage pipelines.
    \item We demonstrate \textbf{Real-Follow-Sim}: the simulation-ready URDFs produced by \method serve as faithful digital twins, enabling zero-shot transfer of manipulation policies trained purely in simulation to real robots.
    \item Extensive experiments demonstrate state-of-the-art performance in geometry, joint accuracy, and physical executability, while running considerably faster than prior multi-stage methods.
\end{enumerate}

\section{Related Work}

\paragraph{3D Generation Method.} Researchers have developed strong geometric priors for synthesizing shapes from text, images, or multi-view observations, enabling scalable reconstruction and synthesis of 3D content.
Single-image-to-3D generation addresses the challenge of recovering full 3D geometry from sparse visual input~\cite{DBLP:journals/corr/abs-2404-07191,triposghighfidelity3dshape,hong2024lrm,zhang2024compress3d,liu2023one,liu2024one,liu2023zero,liu2024syncdreamer}.
Beyond optimization-based pipelines, generative 3D models explore alternative representations and latent structures, including implicit generative models such as Shap·E~\cite{jun2023shap} and GET3D~\cite{gao2022get3d}, as well as structured latent formulations exemplified by Trellis~\cite{trellis, trellis2} and CLAY~\cite{zhang2024clay}.
Gaussian-based 3D generation further improves efficiency and rendering quality through explicit point-based representations~\cite{tang2024dreamgaussian,yi2024gaussiandreamer}.
Complementary to holistic shape synthesis, part-level 3D generation explicitly models semantic parts to improve controllability and compositionality, through either diffusion-based multi-view part reconstruction or autoregressive part generation and discovery~\cite{chen2025partgen,chen2025autopartgen}.

\paragraph{Articulated Object Models.} Currently, they focus on recovering part decomposition, kinematic structure, and motion constraints of human-made objects from visual observations, forming a foundation for robotic interaction and simulation~\cite{liu2025survey}.
Early work primarily studied articulated structure inference from 3D geometry or partial observations, supported by part-level benchmarks and interactive simulation platforms such as PartNet~\cite{mo2019partnet} and SAPIEN~\cite{xiang2020sapien}.
These efforts explored generalizable kinematic modeling, movable or openable part detection, part-level pose estimation, and articulated shape representation, including self-supervised category-level articulation learning and disentangled implicit formulations~\cite{abbatematteo20a,jiang2022opd,liu2023paris,liu2023selfsupervised,mu2021sdf}.
Subsequent approaches shift toward vision-conditioned and controllable modeling, leveraging visual input to recover articulated structure directly from images, including neural implicit representations for articulated digital twins and vision- or vision-language-based articulation reasoning~\cite{weng2024neural,le2025articulateanything,urdformer,liu2025singapo,liu2024cage}.
Alongside vision-conditioned articulation modeling, growing attention has been directed toward generating simulation-ready articulated assets with executable representations, including code-based reconstruction of articulated objects~\cite{mandi2025realcode}, symbolic URDF construction via multimodal language models~\cite{li2025urdfanything}, and physically grounded asset synthesis compatible with physics engines~\cite{cao2025physx}.

\section{Method}

Given a single RGB image, our goal is to infer a complete articulated object model that explicitly represents part geometries and their kinematic relationships. The output is a valid URDF (Unified Robot Description Format) file that can be directly instantiated in standard physics simulators, enabling downstream robot policy learning without manual asset engineering.

Starting from the input image, \method first reconstructs a complete 3D model at the object-level and then autoregressively decomposes it into rigid links together with their articulation parameters. Specifically, we extract visual features using a pretrained DINOv3 encoder~\cite{dinov3} and reconstruct the full object geometry using TripoSG~\cite{triposghighfidelity3dshape}. Based on these global visual and geometric cues, our model sequentially generates part-level geometries and their associated joint specifications, including joint type, axis, origin, and limits. The predicted components are finally assembled into a fully specified URDF model.

\subsection{Task Definition}

Let $\mathcal{X}_{\text{whole}}$ denote the complete object-level 3D geometry capturing the external appearance of an articulated object, typically represented as a single mesh. Importantly, $\mathcal{X}_{\text{whole}}$ does not necessarily coincide with the union of its part-level geometries, as articulated parts may contain internal or functional structures that are not directly observable from the exterior surface.

An articulated object is represented by
\begin{equation}
    \mathcal{A} =
\bigl( \{\mathcal{X}^{(k)}\}_{k=1}^{K}, \{\mathcal{J}^{(k)}\}_{k=1}^{K} \bigr),
\end{equation}
where $\mathcal{X}^{(k)}$ denotes the mesh of the $k$-th rigid link and $\mathcal{J}^{(k)}$ specifies the joint parameters defining the kinematic relationship between part $\mathcal{X}^{(k)}$ and its parent. Each $\mathcal{J}^{(k)}$ includes the joint origin, axis, type, and motion limits. Together, these components define a URDF model that can be directly instantiated in a physics simulator.

We consider the image-to-articulated-model setting: given a single RGB image $I$, the task is to infer a plausible articulated representation $\mathcal{A}$, i.e.\ $I \rightarrow \mathcal{A}$.
Since recovering articulated structure from visual input is inherently ambiguous, we cast it as sampling from a conditional distribution $p(\mathcal{A} \mid I)$, from which valid URDF models can be sampled for downstream real-to-sim-to-real policy learning. We instantiate this distribution through an autoregressive factorization over parts, detailed in the following subsections.
\begin{figure*}[t]
  \centering
  \begin{subfigure}[b]{0.759\textwidth}
    \centering
    \includegraphics[width=\linewidth]{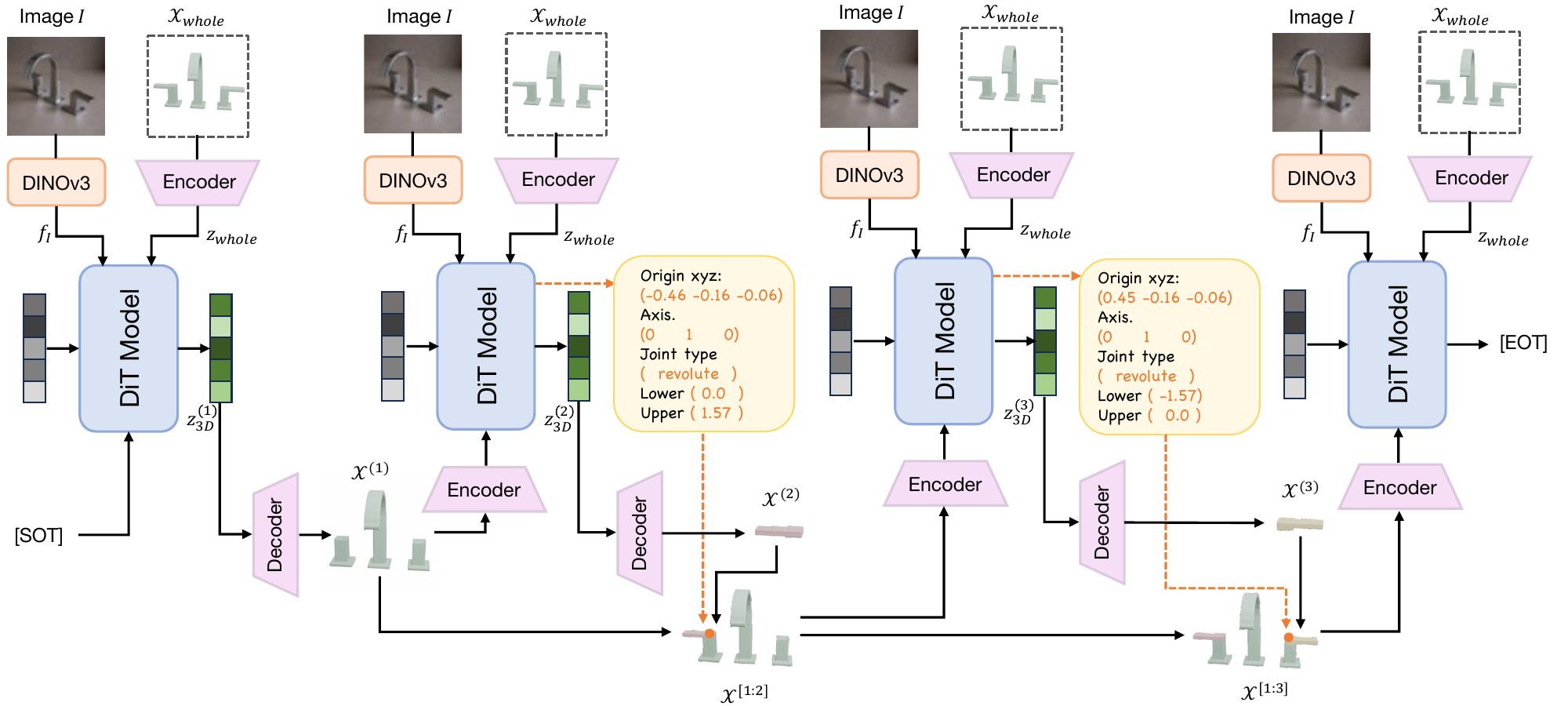}
    \caption{Generation pipeline.}
    \label{fig:model_overall}
  \end{subfigure}
  \hfill
  \begin{subfigure}[b]{0.235\textwidth}
    \centering
    \includegraphics[width=\linewidth]{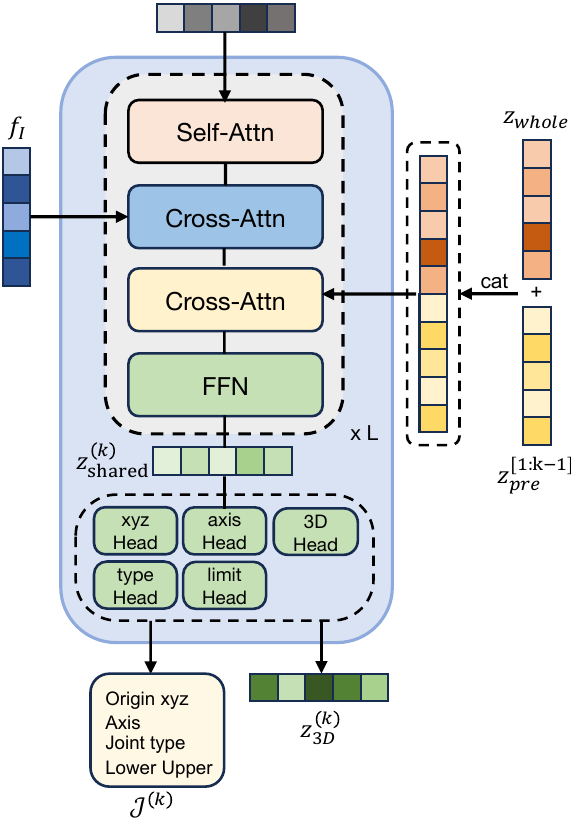}
    \caption{DiT and decoders.}
    \label{fig:dit}
  \end{subfigure}
  \caption{\textbf{Overview of \method.} (a) The autoregressive generation pipeline that produces 3D parts together with their joint specifications and assembles them into a complete URDF. (b) Structure of the DiT backbone and the geometry / articulation parameter decoders.}
  \label{fig:method_overview}
\end{figure*}

\subsection{Latent 3D Shape Modeling via Diffusion}
\label{sec:latent_diffusion}

Directly modeling probability distributions over continuous 3D surfaces is intractable due to their infinite-dimensional nature. We therefore adopt a finite-dimensional latent representation and perform generative modeling in this structured latent space.

We represent 3D geometry with the pretrained encoder-decoder pair $(E, D)$ of TripoSG~\cite{triposghighfidelity3dshape}, which instantiates the \emph{VecSet} formulation~\cite{3dshape2vecset3dshaperepresentation}. Given a surface $\mathcal{X}$, the encoder maps an $N$-point cloud $\mathcal{P} = \left\{ \boldsymbol{p}_i \in \mathbb{R}^3 \right\}_{i=1}^N \subseteq \mathcal{X}$ to a compact latent representation $\boldsymbol{z} = E(\mathcal{P}) = \left( z_1, \dots, z_M \right) \in \mathbb{R}^{M \times d_z}$, where $M \ll N$ and each $z_i$ is a $d_z$-dimensional feature vector. The encoder $E$ is implemented as a permutation-equivariant transformer that subsamples points and aggregates global context via cross-attention. 
The decoder $D$ evaluates the signed distance function by mapping a query point $\boldsymbol{p} \in \mathbb{R}^3$ and the latent tokens $\z$ to
$
\operatorname{SDF}(\boldsymbol{p} \mid \mathcal{X}) = D(\boldsymbol{p} \mid \z),
$
using a Perceiver-style architecture~\cite{perceiveriogeneralarchitecture}. This provides a continuous implicit representation of the 3D shape.

To generate shapes conditioned on visual evidence $y$ (e.g., an input image), we learn a conditional distribution $p(\z \mid y)$ using latent diffusion. We define a forward noising process
\begin{equation}
    \z_t = \sqrt{\bar\alpha_t}\,\z_0 + \sqrt{1-\bar\alpha_t}\,\epsilon,
\quad \epsilon \sim \mathcal{N}(0,I),
\end{equation}
with a predefined noise schedule $\{\alpha_t\}_{t=0}^T$ and $\bar\alpha_t = \prod_{s=1}^{t} \alpha_s$. Following~\cite{salimans2022progressive}, we parameterize the diffusion process via the flow velocity
\begin{equation}
    \bv(t,\z_0,\epsilon)
= \sqrt{\bar\alpha_t}\,\epsilon - \sqrt{1-\bar\alpha_t}\,\z_0.
\end{equation}
A neural network $\hat \bv_\theta(z_t, t \mid y)$ is trained to predict the
velocity at diffusion step $t$, conditioned on the noised latent $z_t$
and the input evidence $y$.
\begin{equation}
\label{loss_diffusion}
\mathcal{L}_{diff} =
\mathbb{E}_{y,\z_0,t,\epsilon}
\bigl\|
\hat{\bv_\theta}(t,\z_t \mid y) - \bv(t,\z_0,\epsilon)
\bigr\|^2.
\end{equation}
After training, the model can sample latent shape representations consistent with the input evidence, which are then decoded into 3D geometry.

\subsection{Autoregressive Articulated Model Generation}
\label{sec:autoregressive}

Our articulated model generation follows the compositional autoregressive paradigm introduced in AutoPartGen~\cite{chen2025autopartgen}, while extending it to jointly synthesize part geometry and articulation parameters within a unified diffusion-based framework.
Fig.~\ref{fig:model_overall} provides an overview of the generation process.

Let $f_I$ denote the DINOv3 image features and $\mathcal{X}_{\text{whole}}$ the complete object-level mesh, which may be either pre-given or produced by an off-the-shelf 3D reconstruction model, e.g., TripoSG~\cite{triposghighfidelity3dshape}. The reconstructed mesh is encoded into a global 3D latent $\z_{\text{whole}} = E(\mathcal{X}_{\text{whole}})$. The pair $(f_I, \z_{\text{whole}})$ serves as global context for autoregressive generation.
At each autoregressive step $k$, the model generates a single articulated part represented by the tuple
$
\bigl( \z^{(k)}_{\text{3D}}, \mathcal{J}^{(k)} \bigr),
$
where $\z^{(k)}_{\text{3D}}$ is a latent code representing the geometry of the $k$-th part and
$\mathcal{J}^{(k)}$ denotes its articulation parameters.
Given the global context $(f_I, \z_{\text{whole}})$ and the context accumulated from previously generated parts $\z^{[1:k-1]}_{\text{pre}}$ (defined in Eq.~\ref{z_pre_definition}),
the joint distribution factorizes autoregressively:
\begin{equation}
p\!\left(
\{\z^{(k)}_{\text{3D}}, \mathcal{J}^{(k)}\}_{k=1}^{K} \mid f_I, \z_{\text{whole}}
\right)
=
\prod_{k=1}^{K}
p\!\left(
\z^{(k)}_{\text{3D}}, \mathcal{J}^{(k)} \mid f_I, \z_{\text{whole}}, \z^{[1:k-1]}_{\text{pre}}
\right).
\end{equation}
The number of parts $K$ is not predetermined: alongside the geometry and articulation targets, the model is trained to predict an end-of-token (\texttt{[EoT]}) latent that signals termination of the autoregression (Appendix~\ref{app:inference}).

\textbf{Latent Diffusion with Shared Latent.}
Part generation is driven by a diffusion transformer (DiT) operating in latent space.
At step $k$, the DiT takes a noisy latent $\z_{t}$ together with conditioning tokens derived from $(f_I, \z_{\text{whole}})$ and $\z^{[1:k-1]}_{\text{pre}}$,
and predicts a denoising velocity field.
Running the reverse process from Gaussian noise yields the \emph{shared latent}
\begin{equation}
\z^{(k)}_{\text{shared}}
\sim
p_\theta\!\left(
\,\cdot\, \mid f_I, \z_{\text{whole}}, \z^{[1:k-1]}_{\text{pre}}
\right),
\end{equation}
which aggregates global geometric structure and contextual information for the current part. Both the geometry head and all articulation heads decode from this single $\z^{(k)}_{\text{shared}}$, hence \emph{shared}.

\textbf{Geometry Latent and 3D Decoding.}
The geometry latent of the $k$-th part is obtained by projecting the shared latent through a dedicated 3D head:
$
\z^{(k)}_{\text{3D}}
=
h_{\text{3D}}\!\left(
\z^{(k)}_{\text{shared}}
\right).
$
We then reconstruct the part geometry from the SDF produced by the TripoSG decoder $D$ via marching cubes:
$
\mathcal{X}^{(k)} = \operatorname{MC}\!\left( D\!\left(\,\cdot\, \mid \z^{(k)}_{\text{3D}}\right) \right).
$

\textbf{Articulation Parameter Decoding.}
Articulation parameters are predicted directly (without diffusion) from the shared latent.
We employ multiple lightweight MLP heads, each responsible for a specific joint attribute:
\begin{align}
\mathbf{o}^{(k)} &= h_{\text{orig}}\!\left(\z^{(k)}_{\text{shared}}\right), \\
\mathbf{a}^{(k)} &= h_{\text{axis}}\!\left(\z^{(k)}_{\text{shared}}\right), \\
\ell^{(k)}_{\text{low}}, \ell^{(k)}_{\text{up}} &= h_{\text{low, up}}\!\left(\z^{(k)}_{\text{shared}}\right), \\
\tau^{(k)}       &= h_{\text{type}}\!\left(\z^{(k)}_{\text{shared}}\right), 
\end{align}
We denote the complete articulation specification as
$
\mathcal{J}^{(k)} =
\bigl(
\mathbf{o}^{(k)},\,
\mathbf{a}^{(k)},\,
\tau^{(k)},\,
\ell^{(k)}_{\text{low}},\,
\ell^{(k)}_{\text{up}}
\bigr).
$
Conditioning articulation prediction on the shared latent allows joint parameters to leverage global context captured during diffusion, leading to geometrically consistent articulation. Fig.~\ref{fig:dit} shows the DiT and parameter decoder's structure.

\textbf{Re-encoding and Autoregressive Context Update.}
After decoding the $k$-th part, we merge it with previously generated geometry:
$
\mathcal{X}^{[1:k]} = \bigcup_{j=1}^{k} \mathcal{X}^{(j)}.
$
Following AutoPartGen~\cite{chen2025autopartgen}, we uniformly sample an $N$-point cloud $\mathcal{P}^{[1:k]} \subseteq \mathcal{X}^{[1:k]}$ from the merged surface and re-encode it with $E$ (Sec.~\ref{sec:latent_diffusion}) to update the context:
\begin{equation}
\label{z_pre_definition}
\z^{[1:k]}_{\text{pre}}
=
E\!\left(
\mathcal{P}^{[1:k]}
\right).
\end{equation}
\textbf{Overall Perspective.}
By iterating over shared-latent diffusion, geometry decoding, articulation prediction, mesh merging, and re-encoding, the proposed framework incrementally constructs a complete articulated object. This design preserves the compositional strengths of AutoPartGen while enabling coherent, end-to-end generation of URDF-ready articulated models.

\section{Experiments}%
\label{sec:experiments}

We evaluate \method along three axes: geometric quality of generated part meshes (Sec.~\ref{ssec:geom}), joint parameter prediction accuracy (Sec.~\ref{ssec:joint}), and end-to-end physical executability via real-robot deployment (Sec.~\ref{sec:real-world-exp}). We further ablate key design choices in Sec.~\ref{ssec:ablation}.
\begin{figure*}[t]
  \centering
  \includegraphics[width=\textwidth]{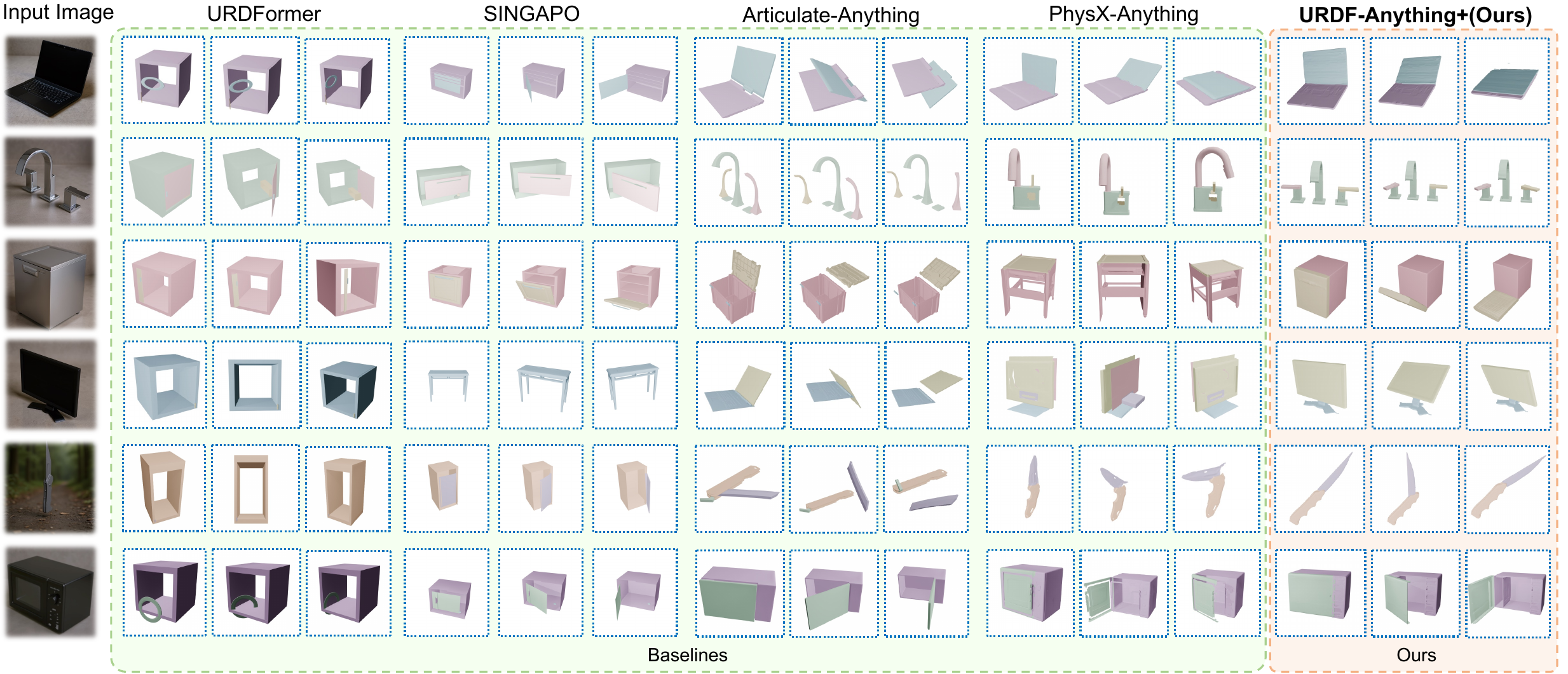}
  \caption{\textbf{Qualitative Results on the Test Set of Our Dataset.} Compared with other methods, URDF-Anything+ generates high-quality 3D assets with more accurate geometry and articulation.}
  \label{fig:visualize1}
\end{figure*}

\subsection{Implementation Details.}





\textbf{Training Details.}
We adopt a two-stage training strategy to stabilize learning and progressively introduce articulation-aware supervision. 
In \textbf{Stage 1}, the model is trained for 500 epochs on 16 NVIDIA A800 GPUs using only image and part-geometry supervision, excluding the end-of-token (\texttt{[EoT]}) signal and URDF-based articulation annotations. This stage focuses on learning robust latent geometry representations and autoregressive part composition. 
In \textbf{Stage 2}, training continues for an additional 100 epochs with the \texttt{[EoT]} signal and URDF-based articulation annotations enabled, thereby introducing full articulation supervision. 
This stage allows the model to jointly learn part termination, articulation parameter prediction, and geometry generation in a unified framework.
More details are reported at Appendix ~\ref{appendix_training} .

\textbf{Training Data.}
Our training data is constructed from a curated collection of articulated 3D datasets, including PartNet~\cite{mo2019partnet}, PartNet-Mobility, and PhysX-3D~\cite{physx3dphysicalgrounded3dasset}.
We apply extensive data cleaning and filtering to ensure geometric validity and consistent articulation structure.
Further implementation details are provided in supplementary material Section \ref{data preparation details} .

\subsection{Geometric Quality Evaluation}
\label{ssec:geom}
\begin{table*}[t]
\centering
\caption{\textbf{Main Results.} Geometry reconstruction quality (Parts and Whole Object), joint parameter prediction accuracy, and average inference time per generated part. We report IoU and F-Score (higher is better, $\uparrow$), CD (lower is better, $\downarrow$), joint Axis (rad), Origin (m), Limit (rad) errors (lower is better, $\downarrow$), and per-part inference time (s, lower is better, $\downarrow$). All reported times include geometry decoding and URDF assembly, and are measured on the same hardware with identical input configurations across methods.}
\label{tab:main_results}
\footnotesize
\setlength{\tabcolsep}{3pt}
\renewcommand{\arraystretch}{1.15}
\resizebox{\textwidth}{!}{%
\begin{tabular}{lcccccccccc}
\toprule
\multirow{2}{*}{\textbf{Method}}
& \multicolumn{3}{c}{\textbf{Parts}}
& \multicolumn{3}{c}{\textbf{Whole Object}}
& \multicolumn{3}{c}{\textbf{Joint Accuracy}}
& \multirow{2}{*}{\makecell[c]{Time / Part\\(s) $\downarrow$}} \\
\cmidrule(lr){2-4} \cmidrule(lr){5-7} \cmidrule(lr){8-10}
& IoU $\uparrow$ & F-Score $\uparrow$ & CD $\downarrow$
& IoU $\uparrow$ & F-Score $\uparrow$ & CD $\downarrow$
& \makecell[c]{Axis Err.\\$\downarrow$} & \makecell[c]{Origin Err.\\$\downarrow$} & \makecell[c]{Limit Err.\\$\downarrow$}
& \\
\midrule
URDFormer~\cite{urdformer}
& 0.202 & 0.389 & 0.354
& 0.243 & 0.413 & 0.286
& \meanstd{1.071}{0.497} & \meanstd{0.504}{0.331} & \meanstd{0.882}{0.679}
&  41.88
\\
Articulate-Anything~\cite{le2025articulateanything}
& 0.563 & 0.452 & 0.064
& 0.627 & 0.510 & 0.053
& \meanstd{0.230}{0.505} & \meanstd{0.267}{0.357} & \meanstd{1.108}{0.676}
&  380.57\\
SINGAPO~\cite{liu2025singapo}
& 0.382 & 0.434 & 0.165
& 0.462 & 0.481 & 0.132
& \meanstd{0.670}{0.472} & \meanstd{0.404}{0.297} & \meanstd{0.817}{0.525}
& 52.30 \\
PhysX-Anything~\cite{cao2025physx}
& 0.784 & 0.697 & 0.076
& 0.813 & 0.715 & 0.068
& \meanstd{0.191}{0.337} & \meanstd{0.103}{0.192} & \meanstd{0.532}{0.423}
& 578.38 \\
\rowcolor{cyan!8}
\textbf{URDF-Anything+ (Ours)}
& \textbf{0.879} & \textbf{0.721} & \textbf{0.033}
& \textbf{0.930} & \textbf{0.742} & \textbf{0.009}
& \textbf{\meanstd{0.129}{0.404}} & \textbf{\meanstd{0.062}{0.096}} & \textbf{\meanstd{0.225}{0.440}}
& \textbf{34.70} \\
\bottomrule
\end{tabular}}
\vspace{-0.5em}
\end{table*}

We compare URDF-Anything+ with the most relevant state-of-the-art methods on articulated 3D geometry generation, including URDFormer~\cite{urdformer}, Articulate-Anything~\cite{le2025articulateanything}, SINGAPO~\cite{liu2025singapo}, and PhysX-Anything~\cite{cao2025physx}.
As shown in Table~\ref{tab:main_results}, URDF-Anything+ consistently achieves superior performance across all geometric metrics, including IoU, F-score, and Chamfer Distance (CD).
In particular, URDF-Anything+ attains the highest IoU and F-score while achieving the lowest CD, indicating more accurate surface reconstruction and better geometric completeness.
Compared to retrieval-based or single-pass generation methods, our approach better captures fine-grained part geometry and preserves global structural consistency across articulated components. We observe that PhysX-Anything and URDF-Anything+ achieve higher performance in geometric generation, as measured by IoU and F-score, which quantify the overlap between the generated geometry and the ground-truth shapes, compared to retrieval-based methods such as URDFormer and Articulate-Anything, which is consistent with our expectations.

\textbf{Qualitative Comparison.}
Fig.~\ref{fig:visualize1} and Fig.~\ref{fig:in-the-wild} compare reconstructions across object categories.
URDFormer~\cite{urdformer} and SINGAPO~\cite{liu2025singapo} rely on a relatively small retrieval library and recover acceptable geometry and joint estimates on simple, box-like categories such as storage furniture and cabinets, but largely fail on shapes that fall outside the library, including faucets, refrigerators, and other objects with curved or thin parts.
Articulate-Anything~\cite{le2025articulateanything} draws from a larger asset base and therefore returns more matching meshes, yet its predicted joint relations are frequently inconsistent—e.g., misaligned axes or origins placed away from the actual hinge—because segmentation and articulation are reasoned about separately.
PhysX-Anything~\cite{cao2025physx} is the strongest baseline overall and yields visually plausible geometry on most categories, but still mispredicts joint axes or part assemblies on objects with thin or repeated structures.
In contrast, \method{} generates parts and joints jointly from a shared latent, producing cleaner part boundaries and more consistent kinematic structure across all shown categories.

\subsection{Joint Parameter Prediction Accuracy.}
\label{ssec:joint}

Following the evaluation protocol of Articulate-Anything~\cite{le2025articulateanything}, we evaluate the accuracy of joint parameter prediction by comparing the predicted URDF parameters with ground-truth annotations.
We focus on articulated objects with \emph{revolute} joints and report errors on three key joint attributes:
(1) \textbf{Joint Axis Error}, defined as the angular difference between the predicted and ground-truth joint axes, normalized to the range $[0,\pi]$;
(2) \textbf{Joint Origin Error}, measured as the Euclidean distance between the predicted and ground-truth joint origins;
and (3) \textbf{Joint Limit Error}, which quantifies the discrepancy between the predicted and ground-truth joint limits.

\begin{figure*}[t]
  \centering
  \includegraphics[width=\textwidth]{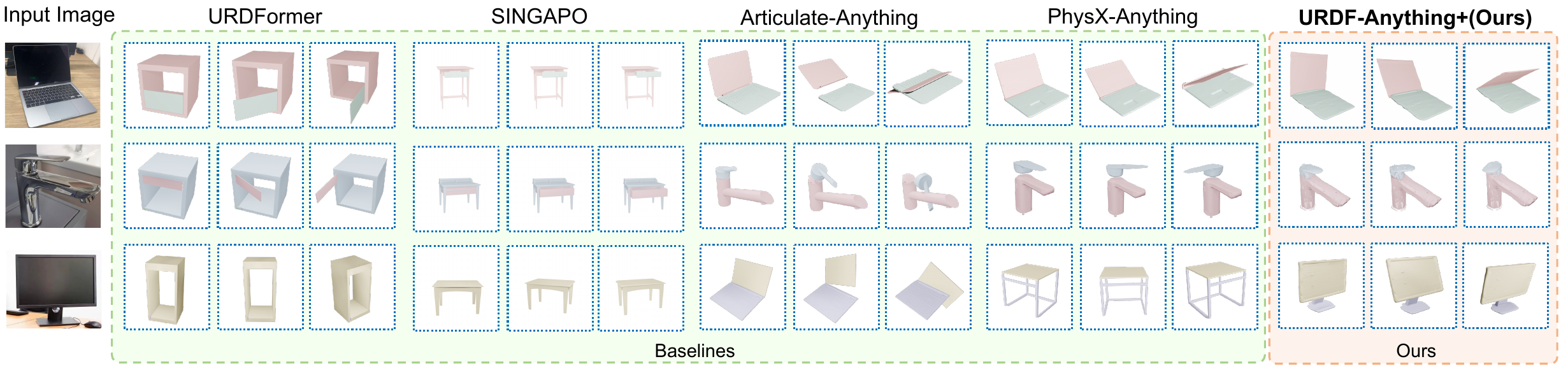}
  \caption{\textbf{Qualitative Results on In-the-wild Images}. Compared with other methods, URDF-Anything+ generates high-quality 3D assets with more accurate geometry and articulation.}
  \label{fig:in-the-wild}
\end{figure*}

Table~\ref{tab:main_results} presents the quantitative results.
We compare URDF-Anything+ (Ours) with several prior methods on the same evaluation set.
As shown in the table, URDF-Anything+ consistently achieves lower errors across all reported joint parameters.
This indicates that our method can more accurately recover the kinematic configuration of articulated objects, particularly in terms of joint geometry and motion constraints.
These results demonstrate the effectiveness of our approach in predicting fine-grained joint parameters.
We attribute the improved performance to our end-to-end framework, which couples autoregressive geometric generation with shared-latent articulation decoding. Beyond the quantitative comparison, we further present qualitative results in Fig.~\ref{fig:visualize1} and Fig.~\ref{fig:in-the-wild}.

\textbf{Inference Efficiency.}
Beyond reconstruction accuracy, we further evaluate the runtime cost of articulated object generation. The rightmost column of Table~\ref{tab:main_results} reports the average inference time across all methods. We measure the wall-clock time elapsed from the moment model loading completes (i.e., all weights and pretrained components are resident in GPU memory) until the autoregressive generation terminates and the final URDF model is produced.
All methods are evaluated on a single NVIDIA A800 GPU with identical input configurations to ensure a fair comparison.  As shown in the table, \method{} achieves comparable or lower inference latency than previous multi-stage pipelines. Existing methods suffer higher overhead from repeated image API calls and reliance on large vision-language models (VLM), while our single-pass end-to-end design eliminates costly external retrieval and symbolic reasoning.

\subsection{Ablation}
\label{ssec:ablation}

\textbf{Effect of Input Modalities.}
\begin{table*}[t]
\centering
\caption{\textbf{Ablation Studies.} We evaluate the impact of input modalities (top) and link ordering strategies (bottom) on geometry reconstruction and joint prediction accuracy.}
\label{tab:combined_ablation}
\resizebox{\textwidth}{!}{
\begin{tabular}{llccccccccc}
\toprule
\multirow{3}{*}{\textbf{Objectives}} & \multirow{3}{*}{\textbf{Variation}} & \multicolumn{6}{c}{\textbf{Geometry Quality}} & \multicolumn{3}{c}{\textbf{Joint Accuracy}} \\
\cmidrule(lr){3-8} \cmidrule(lr){9-11}
& & \multicolumn{3}{c}{\textbf{Parts}} & \multicolumn{3}{c}{\textbf{Whole Object}} & \multirow{2}{*}{Axis Err. $\downarrow$} & \multirow{2}{*}{Origin Err. $\downarrow$} & \multirow{2}{*}{Limit $\downarrow$} \\
\cmidrule(lr){3-5} \cmidrule(lr){6-8}
& & IoU $\uparrow$ & F-Score $\uparrow$ & CD $\downarrow$ & IoU $\uparrow$ & F-Score $\uparrow$ & CD $\downarrow$ & & & \\
\midrule
\multirow{2}{*}{{Input Modality}} 
& Image Only & 0.627 & 0.560 & 0.106 & 0.682 & 0.634 & 0.098 & 0.349 & 0.197 & 0.540 \\
& \textbf{Image + 3D Guid.(Ours)} & \textbf{0.823} & \textbf{0.740} & \textbf{0.028} & \textbf{0.841} & \textbf{0.766} & \textbf{0.009} & \textbf{0.182} & \textbf{0.091} & \textbf{0.281} \\
\midrule
\multirow{2}{*}{Link Ordering} 
& Random Order & 0.630 & 0.686 & 0.145 & 0.710 & 0.704 & 0.072 & 0.203 & 0.168 & 0.442 \\
& \textbf{Order by Joint XYZ (Ours)} & \textbf{0.805} & \textbf{0.742} & \textbf{0.096} & \textbf{0.816} & \textbf{0.753} & \textbf{0.034} & \textbf{0.120} & \textbf{0.102} & \textbf{0.323} \\
\bottomrule
\end{tabular}}
\end{table*}
To validate our design choices, we study the effect of different input modalities on URDF-Anything+ using the \textbf{\emph{faucet}} category, whose geometry is among the most complex across all object classes.
This choice is motivated by the fact that faucets have complicated structures, making it particularly challenging to generate accurate geometry without explicit 3D whole-object guidance.
All models are trained following the same two-stage protocol as the main setting: pretraining for 500 epochs without \texttt{[EoT]} or URDF supervision, followed by 100 epochs of fine-tuning with \texttt{[EoT]} and URDF parameter prediction enabled.
The architecture and training schedule are kept identical across all variants, with differences only in the available input modalities.
As shown in Table~\ref{tab:combined_ablation}, using image input alone allows the model to infer articulation from visual cues but often results in ambiguities in fine-grained geometry and joint placement.
Combining image and 3D inputs consistently achieves better performance across all evaluated metrics, including geometric quality and joint parameter accuracy.


\textbf{Effect of Link Ordering.}
\label{sec:Effect of Link Ordering.}
We study the impact of link ordering in the training data on articulated model generation
using the refrigerator category, as a large portion of refrigerators consists of multiple articulated links.
Specifically, we compare using a random permutation of links with a deterministic ordering
based on the joint origin coordinates, where links are sorted by the $(x, y, z)$ position
of their associated joints in ascending order. The details are shown in Appendix Section~\ref{Ordering of parts}.
Since our model follows an autoregressive generation paradigm, the ordering of links
implicitly defines the factorization of the joint distribution.
As shown in Table~\ref{tab:combined_ablation}, the proposed spatially consistent ordering
leads to more accurate geometry reconstruction and joint parameter prediction,
indicating that a stable and physically meaningful ordering facilitates learning
dependencies among articulated parts.

\subsection{Real-World Experiment}
\label{sec:real-world-exp}
\begin{figure*}[t]
  \centering
  \begin{subfigure}[t]{0.48\textwidth}
    \centering
    \includegraphics[width=\linewidth]{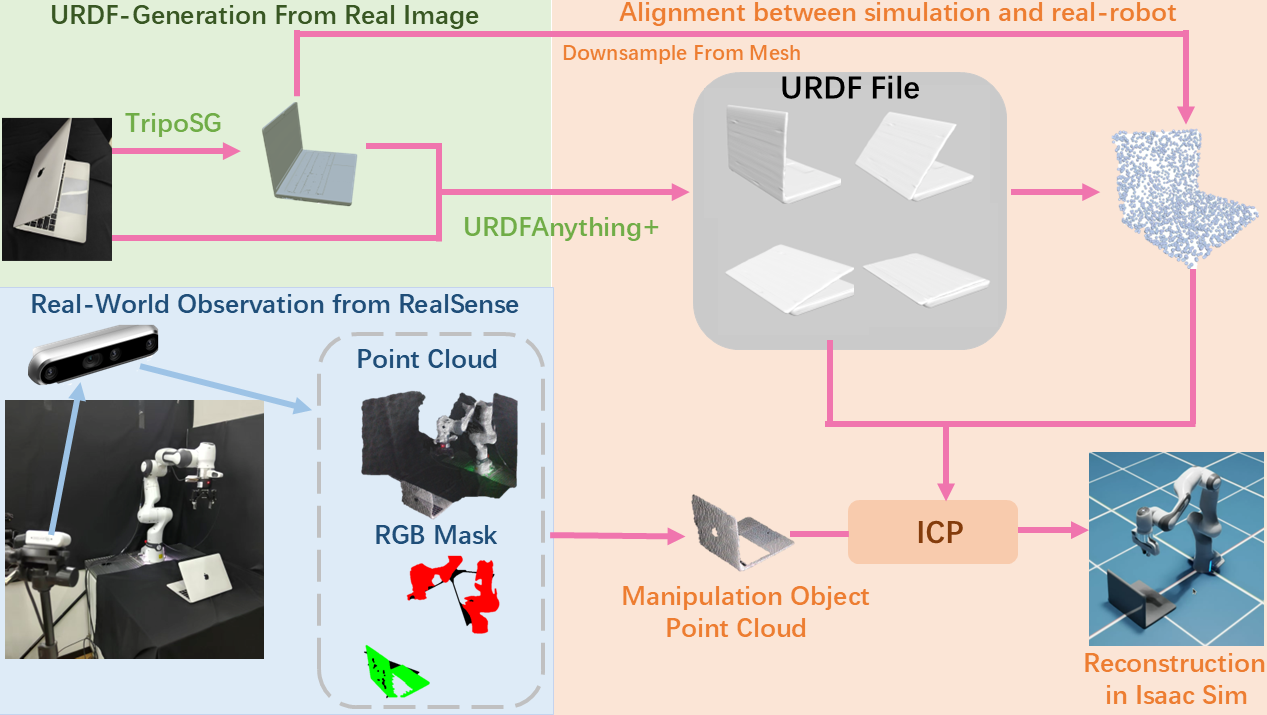}
    \caption{Real-to-Sim digital twin construction: RGB-D capture, \method-generated URDF, ICP alignment, and Isaac Sim instantiation.}
    \label{fig:realfollowsim}
  \end{subfigure}
  \hfill
  \begin{subfigure}[t]{0.48\textwidth}
    \centering
    \includegraphics[width=\linewidth]{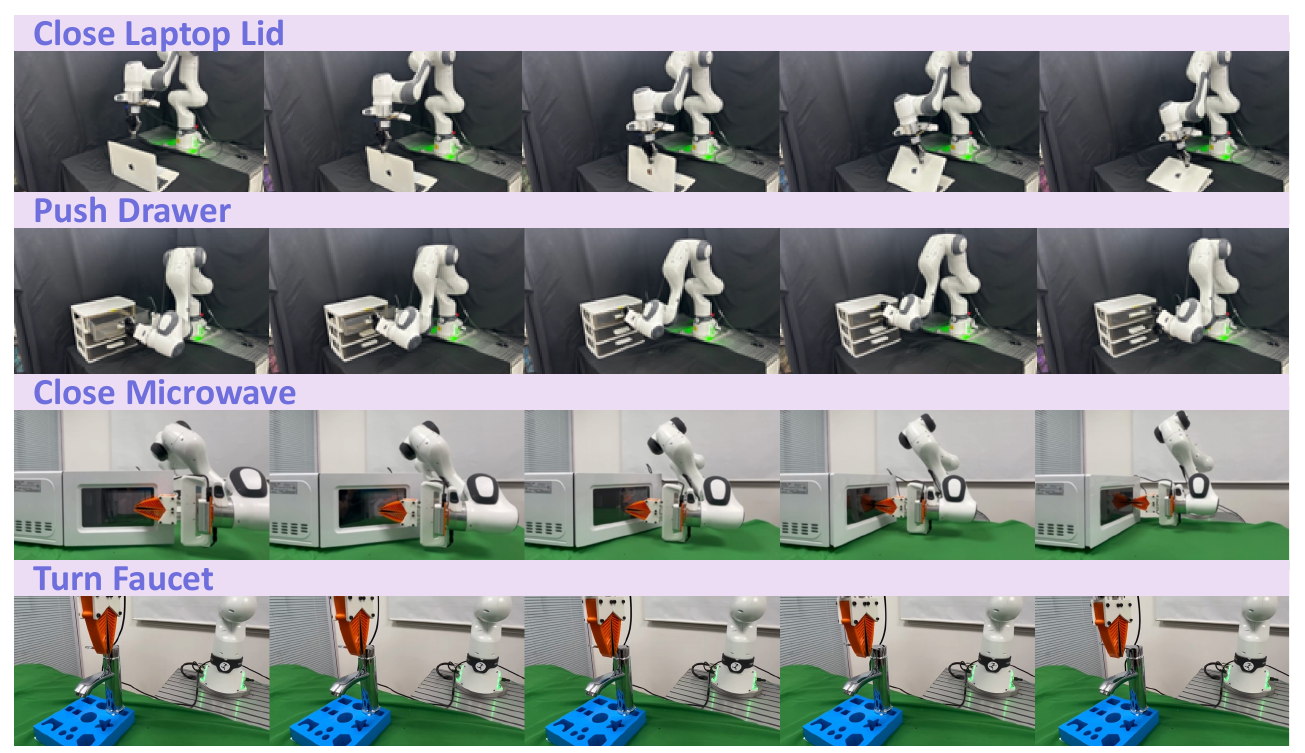}
    \caption{Real-world manipulation tasks used for evaluation: closing a laptop lid and closing a drawer.}
    \label{fig:realtask}
  \end{subfigure}
  \caption{\textbf{Real-Follow-Sim deployment.} (a) Pipeline for constructing the simulation digital twin from real RGB-D observations. (b) The real-world tasks used for evaluation.}
  \vspace{-0.5em}
\end{figure*}

\textbf{Real-Follow-Sim setup.}
Unlike conventional Sim-to-Real approaches that force policies to bridge a fixed sim-real gap, we adopt \emph{Real-Follow-Sim}: the simulator is dynamically aligned to the real scene by streaming reconstructions from \method. As shown in Fig.~\ref{fig:realfollowsim}, we capture RGB-D observations of the scene, generate an articulated URDF via \method, align it with the observed point cloud through ICP~\cite{besl1992method}, scale it to physical size, and instantiate it in Isaac Sim. Policies are trained \emph{exclusively} in this digital twin and executed zero-shot on the real robot, which acts as a faithful follower of simulation trajectories.

We validate the effectiveness of our framework in real-world manipulation tasks using a Franka robotic arm with a Robotiq Gripper~\cite{RobotiqAdaptiveGrippers}. A Realsense camera~\cite{IntelRealSenseL515} is mounted on the side of the robotic arm to provide an ego-centric view of the workspace.
For the real-robot tasks, we evaluate two core manipulation tasks: closing a laptop lid and closing a drawer (see Figure~\ref{fig:realtask}).
In the experiments, we consistently use Diffusion Policy~\cite{chi2025diffusion} for robot policy learning. 
For a fair comparison, both real-world and simulated datasets contain 50 demonstration episodes per task.
Each task is evaluated in real-world scenarios with 20 trials, and task performance is measured by its success rate.

\begin{table*}[t]
    \centering
    \footnotesize
    \setlength{\tabcolsep}{3pt}
    \renewcommand{\arraystretch}{1.15}
    \caption{\textbf{Real-World performance evaluation.} We compare our method across different deployment paradigms, URDF generation baselines, and training data sources. Success rates (\%) are reported.}
    \label{tab:combined_real_world}
    \begin{tabular}{lcc@{\hspace{0.8em}}lcc@{\hspace{0.8em}}lcc}
        \toprule
        \multicolumn{3}{c}{\makecell[c]{\textit{Comparison of}\\\textit{deployment paradigms}}}
        & \multicolumn{3}{c}{\makecell[c]{\textit{Comparison of URDF generation}\\\textit{methods (under Real-Follow-Sim)}}}
        & \multicolumn{3}{c}{\makecell[c]{\textit{Comparison of training}\\\textit{data sources}}} \\
        \cmidrule(lr){1-3} \cmidrule(lr){4-6} \cmidrule(lr){7-9}
        \textbf{Method} & \makecell[c]{\textbf{Close}\\\textbf{Laptop}} & \makecell[c]{\textbf{Close}\\\textbf{Drawer}}
        & \textbf{Method} & \makecell[c]{\textbf{Close}\\\textbf{Laptop}} & \makecell[c]{\textbf{Close}\\\textbf{Drawer}}
        & \textbf{Method} & \makecell[c]{\textbf{Close}\\\textbf{Laptop}} & \makecell[c]{\textbf{Close}\\\textbf{Drawer}} \\
        \midrule
        Sim-to-Real & 80 & 75
        & Articulate-Anything & 75 & 55
        & Real-Only & 85 & 75 \\
        \rowcolor{cyan!8}
        \makecell[l]{\textbf{Real-Follow-Sim}\\\textbf{(Ours)}} & \textbf{100} & \textbf{90}
        & \makecell[l]{\textbf{URDF-Anything+}\\\textbf{(Ours)}} & \textbf{100} & \textbf{90}
        & \makecell[l]{\textbf{Simulation-Only}\\\textbf{(Ours)}} & \textbf{100} & \textbf{90} \\
        \bottomrule
    \end{tabular}
    \vspace{-1em}
\end{table*}

\textbf{Real-Follow-Sim VS. Sim-to-Real.}
We compare our proposed approach, where the physical robot follows motion trajectories generated by the simulated robot, against the conventional paradigm of deploying policies trained in simulation directly onto real-world robots. The  results are shown in Table ~\ref{tab:combined_real_world}, which illustrates that  Real-Follow-Sim paradigm effectively addresses the Sim-to-Real gap by  aligning the simulation with the real world.

\textbf{Comparison with Other Baselines Performance.}
We  compare the real-world performance of different URDF generation methods within our proposed Real-Follow-Sim framework. Table~\ref{tab:combined_real_world} shows that URDF-Anything+   outperforms Articulate-Anything in real-world robot performance.

\textbf{Comparison with Real Data Training.}
In this section, we primarily compare the real-world performance of two policy training paradigms:
(1) policies trained on a dataset collected directly on the real-world robot using a fixed amount of real-world data.
(2) policies trained entirely in simulation using an equally sized dataset generated efficiently under our Real-Follow-Sim framework.
Following the Table \ref{tab:combined_real_world}, we find that our method outperforms URDF generation baselines, improves over the standard Sim-to-Real deployment paradigm, and enables simulation-only training to achieve strong real-world performance.
These results show the effectiveness of Real-Follow-Sim as a unified framework for real-world manipulation.

\section{Conclusion}
We present \textbf{URDF-Anything+}, an end-to-end framework generating physically articulated URDFs from visual observations, outperforming prior methods in geometry, joint accuracy, and physical executability while running considerably faster. Building on this, the \textbf{Real-Follow-Sim} paradigm uses the generated digital twins to enable zero-shot transfer of simulation-trained policies to real robots, pointing toward simulation-ready robotic environments at scale.

\newpage
\bibliographystyle{plain}
\bibliography{ref}


\appendix
\newpage
\section*{Appendix}
This supplementary material provides additional details and results to complement the main paper. It includes the following sections:

\section{Data Preparation Details}
\label{data preparation details}
\subsection{Dataset Expansion Pipeline}
To enhance data diversity, we expanded the \textbf{PartNet-Mobility} dataset by converting static models from \textbf{PartNet}\cite{mo2019partnet} into articulated URDF formats. The pipeline consists of:
\begin{itemize}
    \item \textbf{Automated Synthesis}: We parsed hierarchical JSON metadata and mesh geometries to infer kinematic chains. Contact analysis was performed to determine joint types, transformation matrices, and joint limits.
    \item \textbf{Human-in-the-loop Verification}: All synthesized models were manually inspected to ensure physical plausibility. We specifically corrected joint axes, motion ranges, and collision mesh alignments, while discarding models with topological defects.
\end{itemize}
Meanwhile, the PhysX-3D\cite{physx3dphysicalgrounded3dasset} dataset also provides dynamic URDF annotations generated from PartNet. While the dataset contains a substantial amount of high-quality data, some joint annotations are erroneous. We therefore select a subset of the dataset and manually correct the joint parameters.

\subsection{Normalization}
To ensure a consistent input distribution, all objects underwent the following preprocessing:
We apply standard mesh preprocessing commonly used in 3D geometry pipelines. Each mesh is uniformly scaled from its original resolution while preserving aspect ratios, then translated so that the center of its axis-aligned bounding box is aligned to $(0,0,0)$. After normalization, the mesh is guaranteed to lie within the unit cube $[-1,1]^3$.

\subsection{Mesh Thickening}

\begin{figure}[h]
    \centering
    \begin{subfigure}{0.2\linewidth}
        \centering
        \includegraphics[width=\linewidth, trim=1 1 1 1, clip]{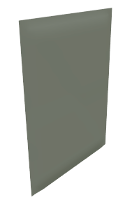}
        \caption{Input mesh}
        \label{fig:single_mesh}
    \end{subfigure}
    \hfill
    \begin{subfigure}{0.2\linewidth}
        \centering
        \includegraphics[width=\linewidth, trim=1 1 1 1, clip]{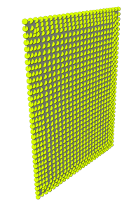}
        \caption{Sampled points}
        \label{fig:single_point}
    \end{subfigure}
    \hfill
    \begin{subfigure}{0.26\linewidth}
        \centering
        \includegraphics[width=\linewidth, trim=1 1 1 1, clip]{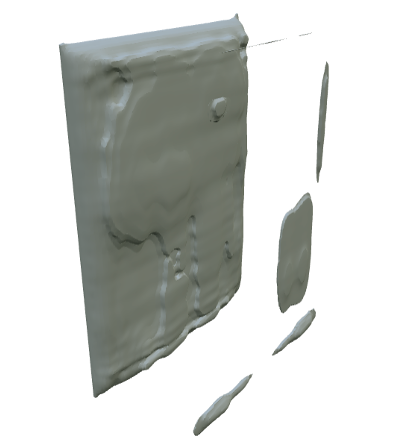}
        \caption{Failed reconstruction}
        \label{fig:single_decode}
    \end{subfigure}
    \caption{Failure case caused by single-layer meshes. From left to right: the input mesh, sampled surface points with unreliable normals, and the failed reconstruction after encoding and decoding.}
    \label{fig:mesh_thickening_motivation}
\end{figure}

Our 3D encoding relies on the encoder from TripoSG\cite{triposghighfidelity3dshape}, which operates on surface geometry and explicitly uses surface normals as input. However, a significant portion of meshes in PartNet and PhysX-3D are single-layer or zero-thickness surfaces. Such meshes often exhibit ill-defined or inconsistent normals, which in practice leads to unstable encoding and frequent reconstruction failures after decoding, which is shown in figure \ref{fig:mesh_thickening_motivation}. 

\begin{figure}[t]
    \centering
    \begin{subfigure}{0.2\linewidth}
        \centering
        \includegraphics[width=\linewidth , trim=1 1 1 1, clip]{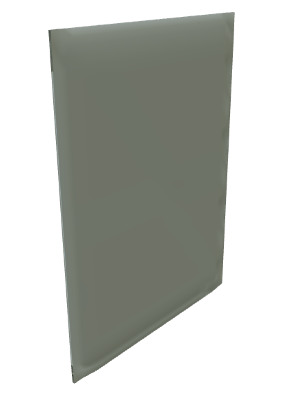}
        \caption{Thickened mesh}
        \label{fig:thick_mesh}
    \end{subfigure}
    \hfill
    \begin{subfigure}{0.2\linewidth}
        \centering
        \includegraphics[width=\linewidth, trim=1 1 1 1, clip]{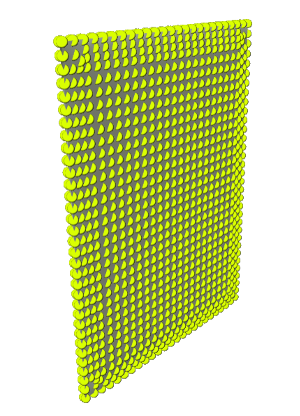}
        \caption{Sampled points}
        \label{fig:thick_point}
    \end{subfigure}
    \hfill
    \begin{subfigure}{0.2\linewidth}
        \centering
        \includegraphics[width=\linewidth, trim=1 1 1 1, clip]{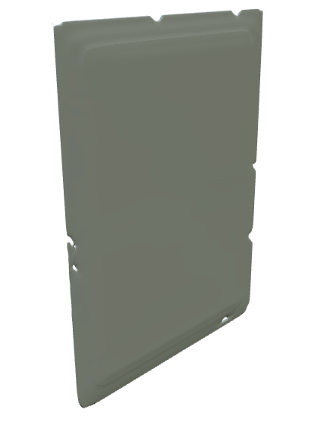}
        \caption{Successful reconstruction}
        \label{fig:thick_decode}
    \end{subfigure}
    \caption{Effect of mesh thickening. From left to right: the thickened mesh, sampled surface points with well-defined normals, and the decoded reconstruction after encoding and decoding. Mesh thickening stabilizes normal estimation and enables reliable latent encoding.}
    \label{fig:mesh_thickening}
\end{figure}

To address this issue, we apply a mesh thickening preprocessing step to affected assets. Specifically, we convert single-layer surfaces into thin volumetric shells by offsetting the mesh along its normal direction, thereby creating a small but non-zero thickness. This process yields well-defined, consistently oriented normals and ensures that the encoder--decoder pipeline can reliably reconstruct the geometry, which is shown in figure \ref{fig:mesh_thickening}. 

As a result, the encoded 3D latents correspond to valid and stable shape representations, which is critical for both diffusion-based generation and subsequent articulated model synthesis.
\begin{figure}[t]
    \centering
    \begin{subfigure}{0.53\linewidth}
        \centering
        \includegraphics[width=\linewidth]{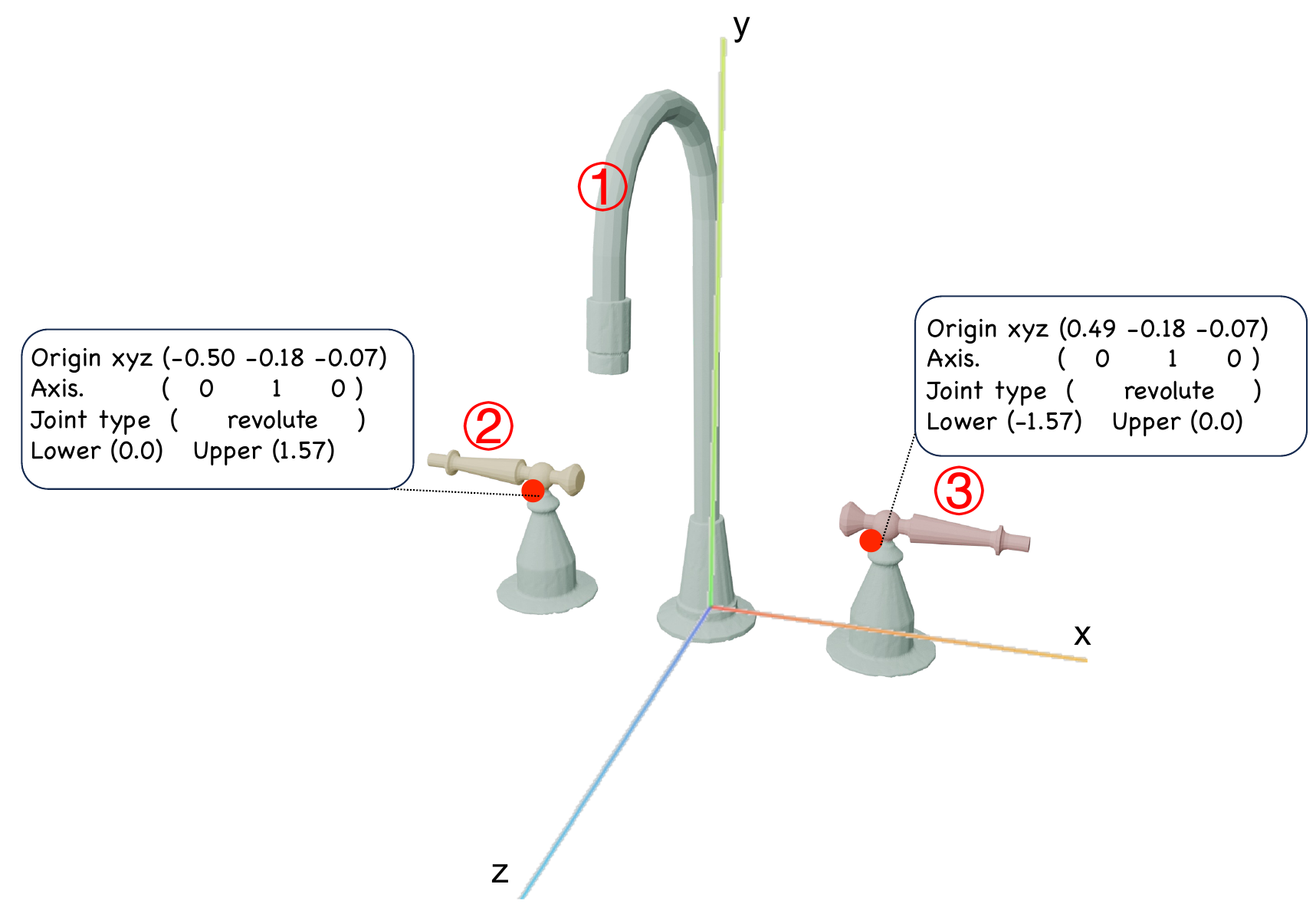}
        \caption{Example 1: Faucet}
        \label{fig:single_mesh}
    \end{subfigure}
    \hfill
    \begin{subfigure}{0.43\linewidth}
        \centering
        \includegraphics[width=\linewidth]{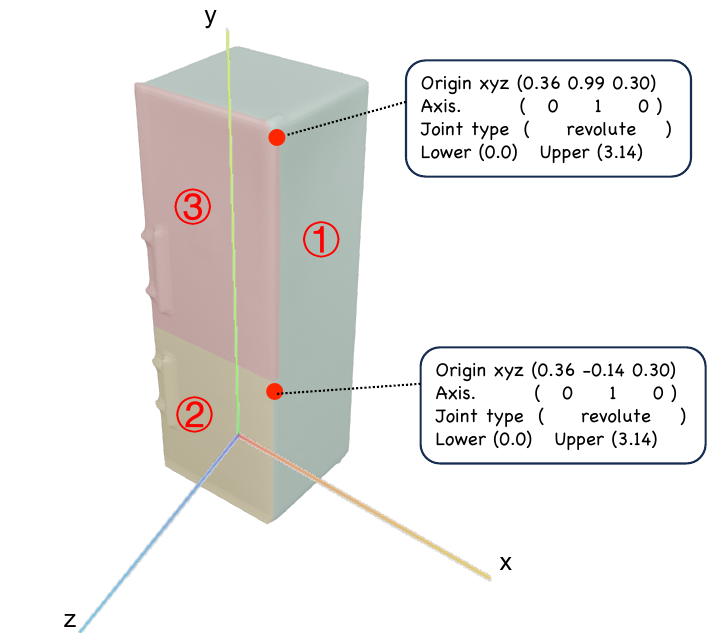}
        \caption{Example 2: Refrigerator}
        \label{fig:single_point}
    \end{subfigure}
    \caption{The visualization of link order setting}
    \label{fig:link_order_visualize}
\end{figure}

\subsection{Ordering of parts}
\label{Ordering of parts}
Our model takes a full image and a complete 3D shape as input and does not impose any explicit constraint on the generation order. To ensure that the model outputs parts in a consistent and semantically correct sequence during autoregressive generation, the training data must follow a predefined ordering scheme.

Specifically, for articulated objects, we designate immobile components such as the base or outer frame as the first part and require the model to generate them first. For the remaining movable parts, we follow the ordering strategy of AutoPartGen\cite{chen2025autopartgen}. 

In the \textit{automatic} setting, a fixed canonical order is used during training. All training assets are defined in a normalized canonical space, and parts are sorted lexicographically based on their axis-aligned bounding boxes: from bottom to top (Z), then left to right (X), and finally front to back (Y), following Blender’s ZXY axis convention. Concretely, parts are first compared by their minimum Z values; if these are similar, minimum X values are compared, and if still similar, minimum Y values are used. The visualization is shown at figure \ref{fig:link_order_visualize}.

The effect of this scheme is discussed in Section \ref{sec:Effect of Link Ordering.}. Here is an example of our data format:
\begin{verbatim}
    {
    "id": "258",
    "whole_image": "base.png",
    "urdf": "test.urdf",
    "links": [
        {
            "name": "link_1",
            "obj": "link_1.obj"
        },
        {
            "name": "link_3",
            "obj": "link_3.obj",
            "origin_xyz": "-0.49765583872795105 -0.17762422561645508 -0.06716569513082504",
            "axis_xyz": "0.0 1.0 0.0",
            "motion_type": "revolute",
            "lower": 0.0,
            "upper": 1.5707963267948966
        },
        {
            "name": "link_2",
            "obj": "link_2.obj",
            "origin_xyz": "0.48748037219047546 -0.1764167696237564 -0.06509745121002197",
            "axis_xyz": "0.0 1.0 0.0",
            "motion_type": "revolute",
            "lower": -1.5707963267948966,
            "upper": 0.0
        }
    ]
}
\end{verbatim}

\subsection{Rendered Images}
\begin{figure}[h]
    \centering
    \begin{subfigure}{0.22\linewidth}
        \centering
        \includegraphics[width=\linewidth]{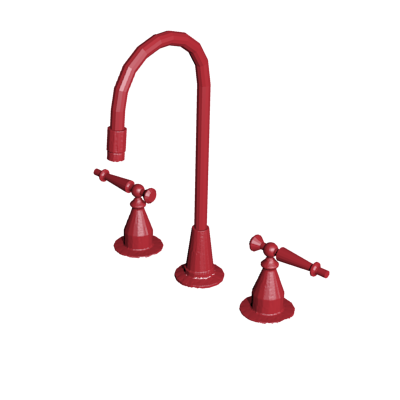}
        \caption{Image from PartNet}
        \label{fig:single_mesh}
    \end{subfigure}
    \hfill
    \begin{subfigure}{0.22\linewidth}
        \centering
        \includegraphics[width=\linewidth]{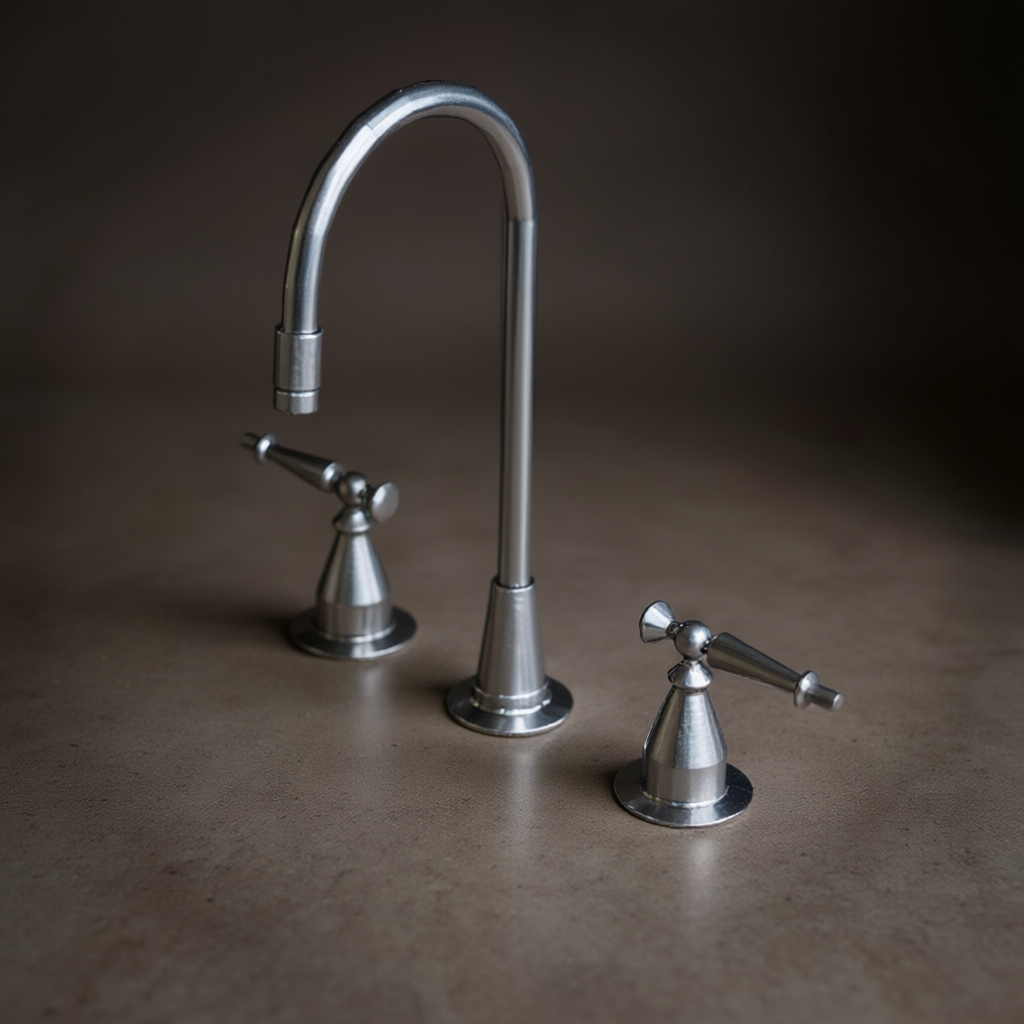}
        \caption{Rendered Image}
        \label{fig:single_point}
    \end{subfigure}
    \hfill
    \begin{subfigure}{0.22\linewidth}
        \centering
        \includegraphics[width=\linewidth]{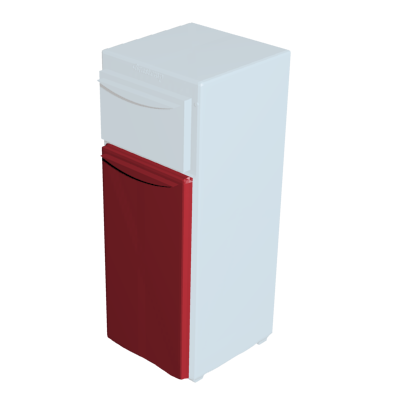}
        \caption{Image from PartNet}
        \label{fig:single_decode}
    \end{subfigure}
    \hfill
    \begin{subfigure}{0.22\linewidth}
        \centering
        \includegraphics[width=\linewidth]{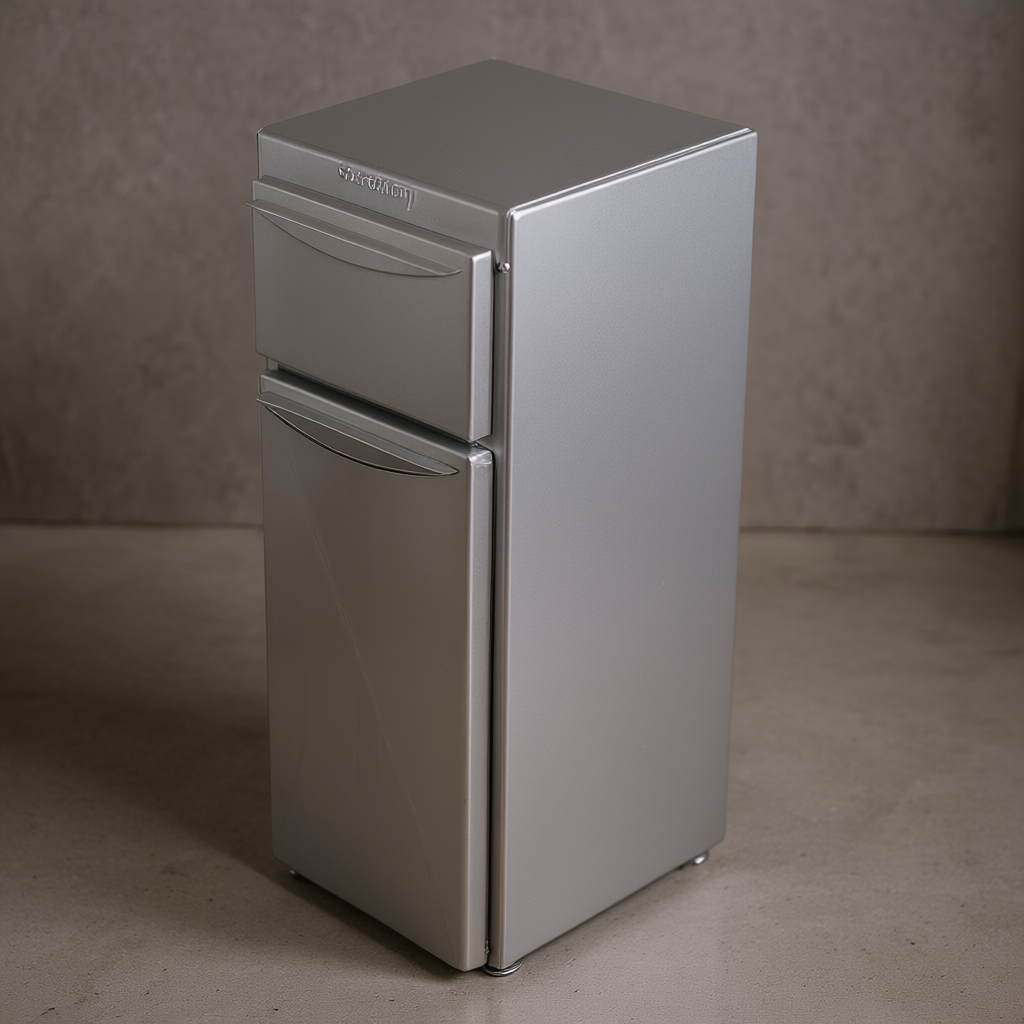}
        \caption{Rendered Image}
        \label{fig:single_decode}
    \end{subfigure}
    \caption{Examples of rendered images generated by Qwen-Image-Edit}
    \label{fig:render_example}
\end{figure}

To improve the generalization of our model to real-world visual inputs, we augment the training data with rendered images. Specifically, we leverage the open-source image editing model \textit{Qwen-Image-Edit} \cite{qwenimagetechnicalreport} to transform object images from the PartNet dataset into images that more closely resemble objects observed in real-world scenes. The examples are shown in figure \ref{fig:render_example}

By enhancing visual realism in terms of lighting, background, and appearance, this rendering process reduces the domain gap between synthetic training data and natural images, thereby improving the robustness of our model when conditioned on input images at inference time.


\section{Implementation Details}
\subsection{Training}
\label{appendix_training}
\paragraph{3D Variational Autoencoder.}
We directly adopt the pretrained 3D VAE from TripoSG~\cite{triposghighfidelity3dshape} without further training. The VAE encodes geometry using a signed distance function (SDF) representation and operates on dense surface samples. During encoding and decoding, we uniformly sample 204{,}800 points per shape and represent each shape using a fixed token length of 512. This pretrained VAE provides stable and high-quality 3D latent representations, which are used throughout diffusion training and inference.

\paragraph{Pretraining and fine-tuning.}
For diffusion training, we follow a two-stage strategy. We first train an image-to-3D diffusion model without URDF parameters or End-of-Token (\texttt{[EoT]}) generation. The diffusion backbone follows DiT~\cite{scalablediffusionmodelstransformers}, consisting of 24 transformer layers with a hidden dimension of 2048. Image conditions are encoded using the DINO-v3 ViT-H/16+ model pretrained on LVD-1689M (\texttt{dinov3-vith16plus-pretrain-lvd1689m}), which has approximately 0.8B parameters. The geometry condition is provided by the pretrained TripoSG 3D VAE.
To reduce computational cost during diffusion training, we precompute the 3D latent representations for all object parts and whole objects (denoted as $\z^{[1:k]}_{\text{pre}}$ and $\z_{\text{whole}}$) by running the 3D VAE offline, and cache the resulting latents  for efficient reuse. This stage is trained for 500 epochs on 16 NVIDIA A800 GPUs.

We then fine-tune the model for an additional 100 epochs by incorporating URDF parameters and \texttt{[EoT]} generation into the training objective. Unless otherwise specified, all other training settings, optimization strategies, and diffusion configurations remain identical to those used in the initial training stage.

\paragraph{Training loss configuration and hyperparameter settings}
During the pretraining stage, the diffusion model is optimized solely using the latent diffusion objective $\mathcal{L}_{\text{diff}}$ defined in Eq.~\ref{loss_diffusion}, which supervises the denoising velocity prediction in latent space. In the fine-tuning stage, we extend the training objective to jointly supervise articulated structure generation by incorporating additional losses. Specifically, the total loss is given by
\[
\mathcal{L}=
\mathcal{L}_{\text{diff}} +
0.1 * \mathcal{L}_{\text{EOT}} +
 0.01*\mathcal{L}_{\text{joint}},
\]
where $\mathcal{L}_{\text{EOT}}$ denotes the End-of-Token (\texttt{[EoT]}) prediction loss, and $\mathcal{L}_{\text{joint}}$ aggregates losses for joint parameter prediction. For continuous joint parameters (e.g. joint origin, axis direction, and limits), we use $\ell_1$ loss, while joint type prediction is supervised using a categorical cross-entropy loss. 

Notably, for a large portion of articulated objects, the joint rotation axes are aligned with canonical directions (e.g., $(0,0,1)$, $(0,1,0)$, or $(1,0,0)$). In such cases, the coordinate of the joint origin along the axis direction does not affect the kinematic behavior of the joint. To avoid introducing unnecessary noise into training, we mask out the origin component along the predicted joint axis when computing the origin $\ell_1$ loss. This axis-aware masking focuses supervision on the physically meaningful degrees of freedom and leads to more stable and accurate joint parameter learning.

All models are optimized using AdamW~\cite{adamw} with a learning rate of $1\times10^{-5}$ during fine-tuning, while other optimization settings follow those used in pretraining.

\subsection{Inference}
\label{app:inference}
At inference time, the articulated object is generated in an autoregressive manner until a termination condition is met. We explicitly introduce an End-of-Token (\texttt{[EoT]}) signal in the latent space to indicate the end of part generation. Specifically, the \texttt{[EoT]} target is defined as a zero vector with the same dimensionality as the shape latent, i.e., a zero latent that matches the geometry latent representation.

During inference, after each autoregressive step, we measure the distance between the predicted latent and the \texttt{[EoT]} latent. If this distance falls below a predefined threshold, the generation process is terminated, indicating that no further valid parts remain to be generated.

In addition, we incorporate a geometric redundancy check to prevent duplicate or degenerate part generation. After decoding the current part geometry, we evaluate its overlap and similarity with the previously generated part. If the overlap exceeds a predefined similarity threshold, indicating that the newly generated link is largely redundant with an existing one, the generation is also terminated.

\subsection{Evaluation Metrics}

We use three standard metrics to evaluate the reconstruction quality: Intersection over Union (IoU), F-score, and Chamfer Distance (CD).

\paragraph{Intersection over Union (IoU).} 
IoU measures the volumetric overlap between the predicted shape $P$ and the ground-truth shape $G$:

\begin{equation}
\text{IoU} = \frac{|P \cap G|}{|P \cup G|}
\end{equation}

where $|\cdot|$ denotes the volume (number of occupied voxels or mesh volume) of the shape.

\paragraph{F-score.} 
F-score measures the geometric similarity between two point sets $P$ (predicted) and $G$ (ground truth) based on a distance threshold $\tau$. In our experiments, we set $\tau = 0.02$:

\begin{align}
\text{Precision} &= \frac{|\{ p \in P \mid d(p, G) < \tau \}|}{|P|}, \\
\text{Recall} &= \frac{|\{ g \in G \mid d(g, P) < \tau \}|}{|G|}, \\
\text{F-score} &= \frac{2 \cdot \text{Precision} \cdot \text{Recall}}{\text{Precision} + \text{Recall}},
\end{align}

where $d(p, G) = \min_{g \in G} \| p - g \|_2$ is the distance from a predicted point to the nearest ground-truth point.

\paragraph{Chamfer Distance (CD).} 
Chamfer Distance measures the average closest-point distance between two point sets:

\begin{equation}
\text{CD}(P, G) = \frac{1}{|P|} \sum_{p \in P} \min_{g \in G} \| p - g \|_2^2
+ \frac{1}{|G|} \sum_{g \in G} \min_{p \in P} \| g - p \|_2^2
\end{equation}

This metric captures the overall similarity between two point clouds.

\section{Visualization of Generated Shapes Across Object Categories}
To qualitatively evaluate the performance of our method, we visualize the generated shapes for different object categories. 
\begin{figure*}[h]
  \centering
  \includegraphics[width=0.815\textwidth]{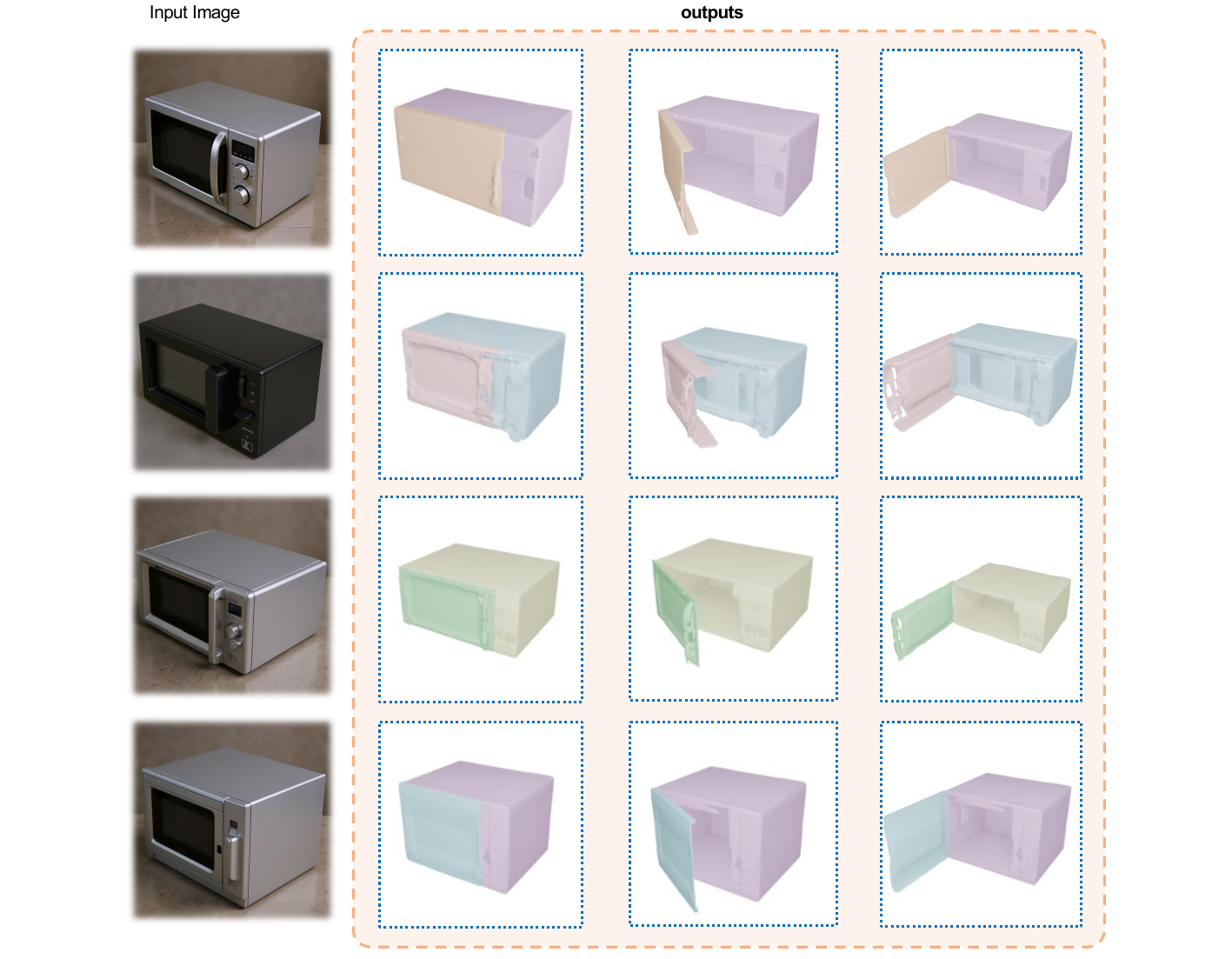}
  \caption{Generation Results of Microwaves.}
 
\end{figure*}
\begin{figure*}[h]
  \centering
  \includegraphics[width=0.7\textwidth]{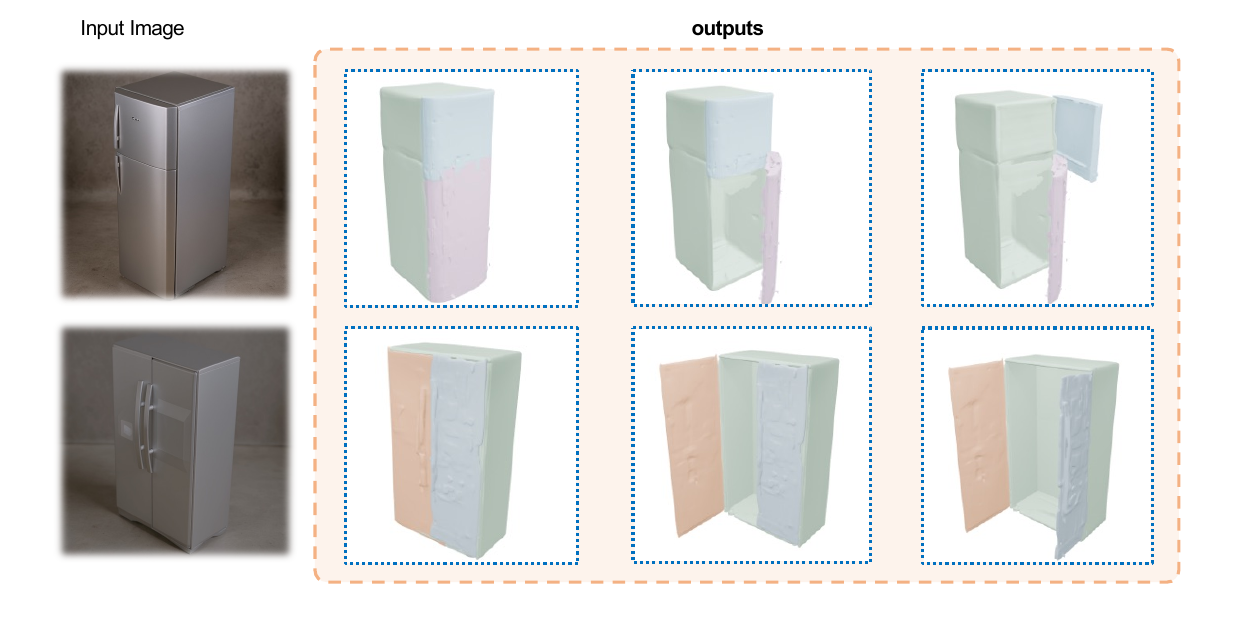}
  \caption{Generation Results of Refrigerators.}
  
\end{figure*}
\begin{figure*}[h]
  \centering
  \includegraphics[width=0.66\textwidth]{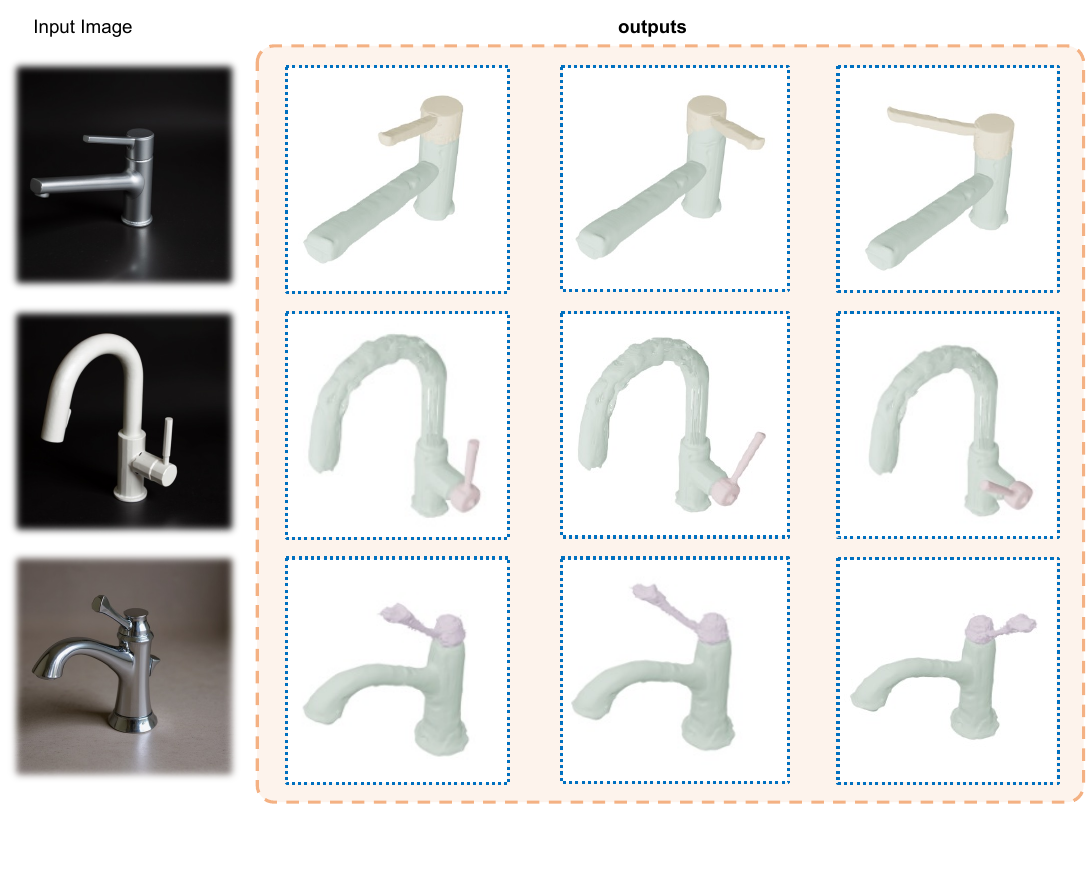}
  \caption{Generation Results of Faucets.}
  
\end{figure*}
\begin{figure*}[h]
  \centering
  \includegraphics[width=0.7\textwidth]{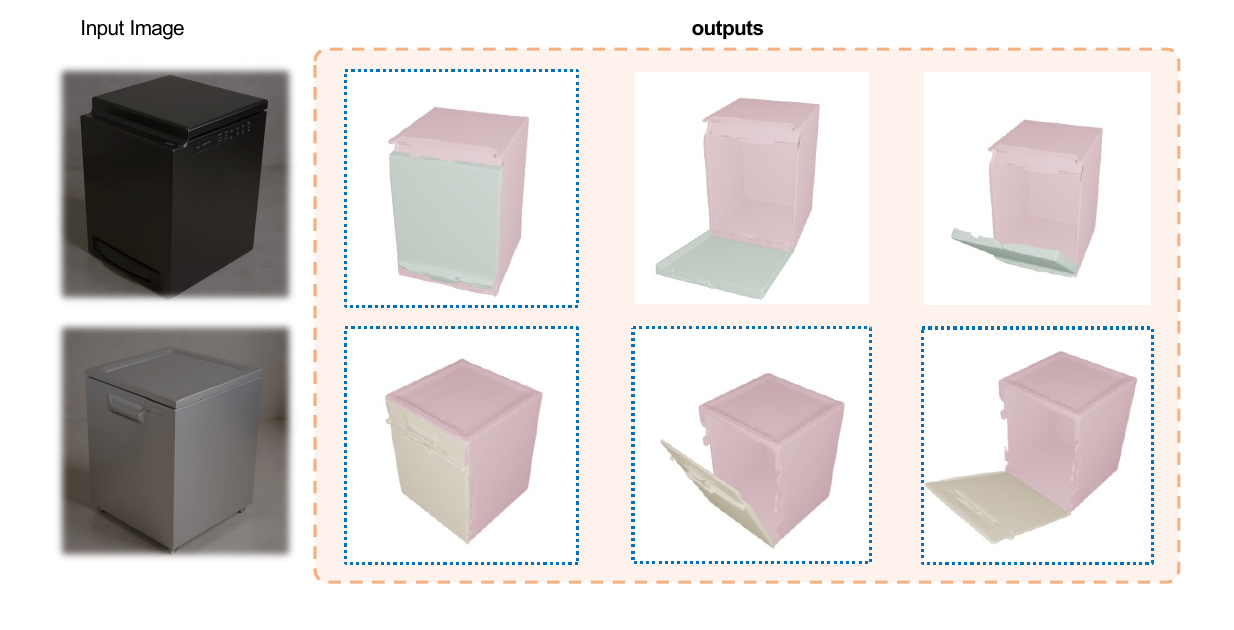}
  \caption{Generation Results of Dishwashers.}
  
\end{figure*}

\section{Limitations}
\label{app:limitation}
While \method achieves strong performance in generating articulated object models and enabling the Real-Follow-Sim pipeline, several limitations remain. First, the diversity of articulated objects in the training data is still limited. Our model primarily relies on publicly available datasets, where certain object categories contain relatively few instances. This data imbalance may restrict the model’s ability to generalize to rare or unseen articulated structures. Expanding the dataset with more diverse articulated objects and richer annotations would likely further improve performance and robustness.

Second, although our method predicts geometric structure and kinematic parameters, it does not explicitly model more complex physical properties of objects, such as mass distribution, friction, or material-dependent dynamics. As a result, some physical behaviors in simulation may not perfectly match those in the real world. Incorporating richer physical parameter estimation or physics-aware learning could further improve the fidelity of the generated digital twins.

\end{document}